# Unleashing the Potential of Large Language Models: A Blueprint for Real-Time, Enterprise-Ready Deployments

Muhammad Faizan Raza[1] · Shuo (Luna) Yang[2] · Satish Mahadevan Srinivasan[1],* · Joanna F. DeFranco[3]

[1] Penn State Great Valley, Malvern, PA 19355, USA [2] Penn State Brandywine, Media, PA 19063, USA [3] The Pennsylvania State University, USA

* Corresponding author: Satish Mahadevan Srinivasan (sus64@psu.edu)

ORCID — Raza: 0009-0005-3256-1130 · Yang: 0000-0003-2390-243X · Srinivasan: 0000-0003-1377-3726

*This article describes a practical "always updating" architecture for large language model pipelines by connecting to streaming data, learning and fact-checking, and automatically improving from errors and human feedback to remain timely, reliable, and fast.*

In today's continuous data and rapidly evolving information environments, large language models (LLMs) have become a transformative technology, enabling new capabilities in health care, finance, and customer support [1]. LLM deployment can be distributed, event-driven pipeline spanning edge devices, gateways, and cloud services. Trained on massive corpora, they support applications from conversational agents to high level synthesis and can match or exceed human benchmarks in tasks such as translation and structured reasoning [2], [3]. As organizations deploy LLMs in real-time settings (for example, financial surveillance or emergency response platforms), critical limitations become more consequential: knowledge staleness from static training, catastrophic forgetting during updates, hallucinations that generate plausible but inaccurate outputs, and weak feedback loops that slow detection and correction. In health care, outdated clinical guidelines can distort clinical decisions; in finance, obsolete market insights can amplify losses. We address these risks by introducing a unified, pattern-driven LLM operations (LLMOps) architecture designed for dynamic, continuously updated operational use.

## THE EVOLUTION OF LLM DEPLOYMENTS AND THEIR INHERENT LIMITATIONS

LLM deployment has progressed in recognizable stages, each amplifying the tension between expanding model capability and the operational demands of real-world use. In the initial phase, represented by models such as BERT (Bidirectional Encoder Representations from Transformers) and early versions of GPT (Generative Pre-trained Transformer), LLMs were treated as static components, fine-tuned for narrowly defined tasks and deployed with no mechanism for ongoing adaptation [2], [4]. Although this approach was effective for early use cases, it relied on the assumption of a stable knowledge base, an assumption that proved unrealistic in rapidly evolving domains. For instance, a model trained on data available up to 2022 might confidently describe COVID-19 protocols while failing to incorporate developments related to 2024 variants, thereby contributing to misinformation in public health guidance.

Models like GPT-3 and PaLM (Pathways Language Model) introduced in-context learning, allowing LLMs to adapt to behaviors via prompts instead of retraining [2], [5]. This advancement improved flexibility but revealed weaknesses with time-sensitive data: limited-context windows restrict access to up-to-the-minute information and leave persistent knowledge gaps. For instance, a financial analyst using these models could receive recommendations based on stale market trends, heightening risk in fast-moving trading environments. The latest, state-of-the-art models such as GPT-5 and Llama-4 exhibit unprecedented scale and capability, yet they also expose significant shortcomings in deployment. A study by Zhu et al. [6] suggested that knowledge staleness can reduce model accuracy by 15%–25% in domains characterized by rapid information change, such as regulatory compliance and news analysis. Additionally, catastrophic forgetting, where fine-tuning processes overwrite previously learned knowledge, has been shown to cause performance declines of 20%–40% on unrelated tasks [7], [8]. Hallucinations, which occur at rates of 15%–30% in knowledge-intensive applications, further erode reliability; for example, an LLM used in legal services may fabricate case precedents, potentially leading to compliance violations [9], [10].

These limitations intensify in enterprise contexts. Regulatory requirements, such as the General Data Protection Regulation (GDPR) in Europe and the Health Insurance Portability and Accountability Act (HIPAA) in the United States, require traceability, auditability, and controlled change management, properties that many ad hoc deployments do not natively provide [11], [12]. Meanwhile, production expectations (for example, high availability and low latency) sit uneasily with the cost and variability of LLM inference, and the overhead governance required to make outputs reviewable and reversible. Recent practitioner analyses emphasize how compliance and control processes can consume a large share of implementation effort, slowing iteration and delaying error correction.

To address these gaps, we propose a comprehensive framework (see Figure 1) that formalizes LLM deployment as an operational discipline, managing tradeoffs among latency, cost, and factuality, while aligning system behavior with organizational controls and regulatory constraints. The result is an LLM pipeline that adapts with its environment and converts deployment risks into governed, monitorable capabilities.

## FRAMEWORK

In this article, we present a framework that integrates real-time data ingestion, continual learning, retrieval-augmented generation (RAG), and human-in-the-loop feedback into a single operational pipeline, thereby transforming LLMs from brittle models into adaptive and dependable enterprise systems (see Figure 1) [13]. It addresses the latency, reliability, and consistency challenges that define networks and connected systems.

By incorporating established software design patterns, the architecture reduces latency–cost–accuracy tradeoffs while meeting governance needs such as auditability and rollback. This framework enables safer deployment of LLMs in high-risk, tightly regulated sectors. Designed for practitioners, it traces the evolution of deployment strategies, tackles challenges in real-time operational contexts, and delivers novel methodological tools, all culminating in a practical, robust architecture that unlocks the full value of LLM technology with confidence and compliance.

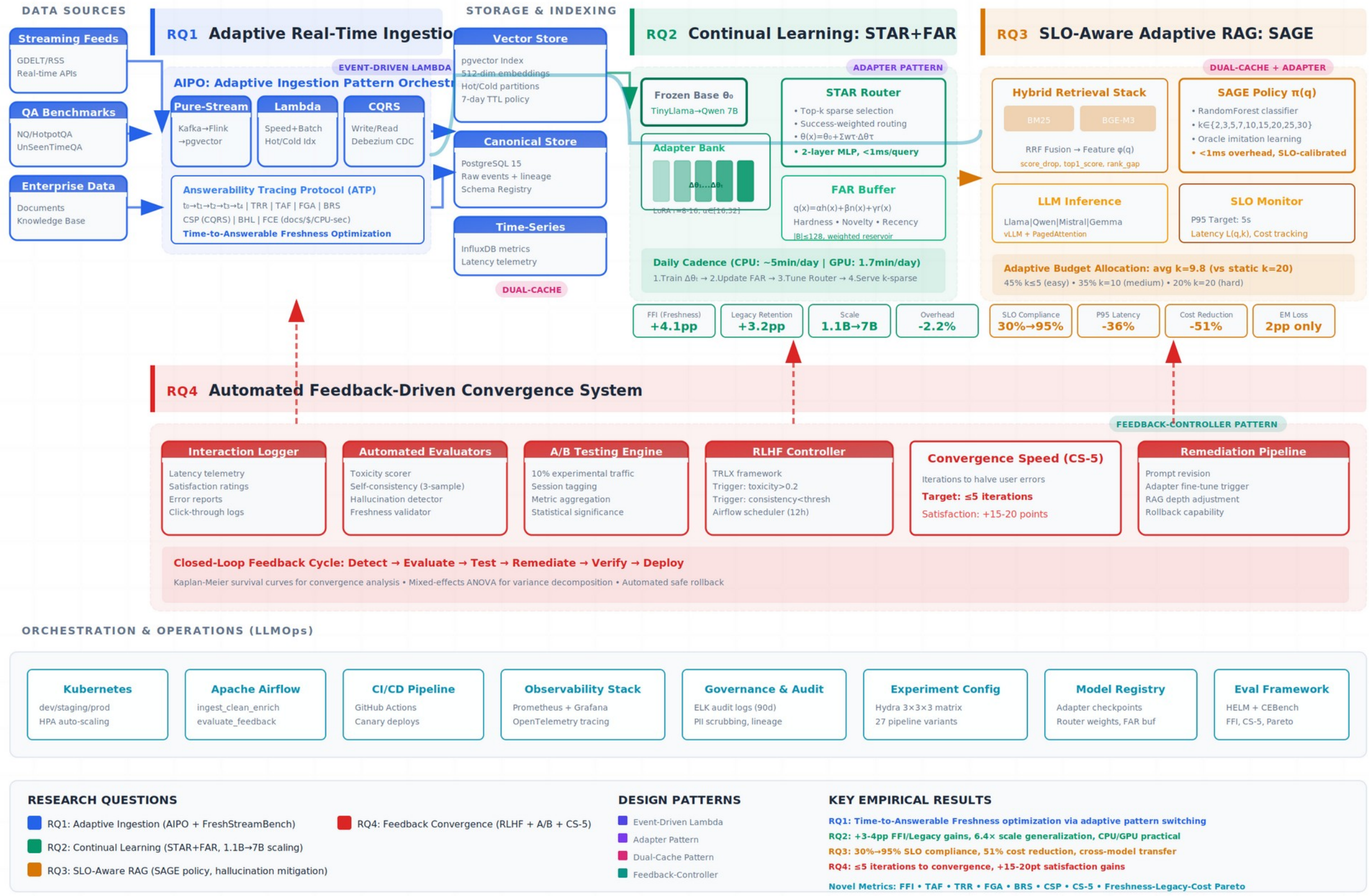

**Figure 1.** Strategic pipeline blueprint for enterprise-ready real-time LLM systems. The architecture integrates four research contributions: (1) an adaptive ingestion pattern orchestrator with FreshStreamBench evaluation framework; (2) STAR+FAR continual learning with sparse temporal routing; (3) SAGE, an SLO-aware adaptive retrieval for hallucination mitigation; and (4) an automated feedback-driven convergence with RLHF triggers. The design patterns (lambda adapter, dual cache, and feedback controller) are mapped to each stage and validated through extensive experiments, producing Pareto-optimal configurations. QA: quality assurance; RQ: research question; STAR+FAR: sparse temporal adapter routing + freshness-aware replay; CQRS: Command Query Responsibility Segregation; MLP: multilayer perceptron; SLO: service-level objective; A/B: alpha/beta; RAG: retrieval-augmented generation; config: configuration; eval: evaluation.

## PROBING THE PILLARS OF LLM PIPELINES

Central to our proposed framework are the four guiding pillars (latency, improve knowledge freshness, limit hallucinations, and accelerating feedback mechanisms), each addressing a key aspect of LLM deployment and collectively demonstrating that pattern-driven architecture can concurrently reduce the pillars without incurring prohibitive costs. These pillars extend existing developments in individual areas, such as RAG and continual learning, by synthesizing them within a unified architectural framework.

### Pillar 1: Freshness Latency

First, regarding real-time data ingestion, our framework evaluates pure-stream, event-driven lambda, and Command Query Responsibility Segregation (CQRS) patterns using FreshStreamBench and the Answerability Tracing Protocol, which measures time-to-answerable freshness from document arrival to the first verified correct, retrieval-grounded answer. In steady workloads, pure-stream processing (Kafka/Flink to a vector read model) minimizes time-to-retrieval-ready but remains on par with other techniques under bursts and backpressure at high load. Lambda architectures pair a speed layer with a batch backfill layer, yielding stronger burst resilience, faster freshness-gap recovery, and predictable throughput across heterogeneous enterprise sources. CQRS separates the write model from the query/read model via projections driven by change data capture from the write store’s transaction log/event stream (row-level inserts/updates/deletes), improving read-path efficiency while quantifying eventual-consistency risk via staleness probability. We further introduce an adaptive ingestion pattern orchestrator, an adaptive controller that switches ingestion modes to dominate static patterns on freshness–latency–cost frontiers.

### Pillar 2: Continual Learning

Second, with respect to continual learning, we investigated how experience replay, low-rank adaptation (LoRA), and hybrid strategies balance factual freshness against catastrophic forgetting over time [8]. Experience replay mitigates knowledge erosion by retaining historical examples but can raise storage and privacy risks in regulated environments. LoRA is parameter-efficient and supports incremental updates, yet LoRA-only updates can forget without stabilization. We proposed a sparse temporal adapter routing + freshness-aware replay (STAR+FAR) framework that combines per-day LoRA adapters with query-conditioned sparse routing and a bounded replay buffer that prioritizes recent/novel/hard examples to adapt under drift while mitigating forgetting. We evaluated the rolling daily update cycles on day-wise streams from Wikipedia revisions, news, and StackExchange to determine the factual freshness index alongside legacy retention and cost. Across seeds and model scales, STAR+FAR yielded three- to four-point gains in freshness and legacy under practical CPU/GPU budgets.

### Pillar 3: Reliability

Third, to address the challenges with respect to hallucinations and responsiveness, we propose the enhancement of RAG under tail-latency service level objectives (SLOs) (for example, a 5-s P95 latency target) and cost budgets [14]. We propose SAGE (SLO-aware Adaptive Grounding Engine), which predicts a per-query passage budget k from the lightweight retrieval features (for example, score distributions, rank gaps, and lexical signals) and is trained offline via imitation learning from an oracle budget sweep, adding no extra LLM calls in production [13]. On natural questions under a 5-s P95 SLO [the system must stay in the 95th-percentile (tail) end-to-end response latency at or below 5 s so that at least 95% of queries finish within 5 s], SAGE improved SLO compliance from 30% (static k=20) to 95%, reduced P95 latency by 36%, and reduced retrieval cost by ~51%, with only a two-point exact-match drop. The same policy transfers to HotpotQA, UnSeenTimeQA, and multiple LLM families without retraining. This generalization holds because SAGE operates on retrieval-stage, model-agnostic features (for example, score/rank statistics and lexical signals) so that the learned mapping from feature patterns to an appropriate passage budget k can be reused across datasets and LLM back ends without updating policy weights [9].

### Pillar 4: Feedback Loops

Finally, we formalize feedback integration, combining automated evaluators, alpha/beta (A/B) testing, and reinforcement learning from human feedback (RLHF) triggers to accelerate detection and correction of failures while maintaining governance control [15]. Automated evaluators discover issues in production, A/B tests validate changes with users, and RLHF is reserved for cases where persistent error patterns warrant preference alignment.

In conclusion, this research advances the field of LLM deployments by introducing a unified LLMOps architecture that transforms LLMs from static, error-prone tools into adaptive, enterprise-ready systems capable of operating in dynamic, real-time environments. By addressing key challenges such as knowledge staleness, catastrophic forgetting, hallucinations, and inadequate feedback loops, our framework integrates real-time data ingestion through AIPO, continual learning through the STAR+FAR method, SLO-aware adaptive retrieval via SAGE, and automated feedback mechanisms to deliver a robust pipeline that balances latency, cost, and accuracy while ensuring auditability and regulatory compliance.

The empirical results underscore the architecture's effectiveness, with evaluations on diverse datasets demonstrating substantial gains. STAR+FAR improved factual freshness while mitigating forgetting, and SAGE achieved 95% SLO compliance with a significant reduction in retrieval costs and minimal accuracy loss. These outcomes not only validate the proposed design patterns, such as lambda adapter and dual cache, but also highlight their potential to foster Pareto-optimal configurations in high-risk sectors like health care and finance, where outdated information could lead to significant consequences.

Ultimately, this blueprint empowers organizations to harness the full potential of LLMs with greater confidence, promoting safer deployments that align with stringent regulations like the GDPR and HIPAA. As LLM technologies continue to evolve, future research could extend this framework by incorporating advanced multimodal data sources or scaling them to larger model families, thereby further enhancing their adaptability and impact. This work represents a critical step toward realizing the transformative value of LLMs in enterprise settings, ensuring that they evolve alongside the world's accelerating data landscape.

## ACKNOWLEDGMENT

Satish Mahadevan Srinivasan is the corresponding author.

## ABOUT THE AUTHORS

**Muhammad Faizan Raza** is a graduate student in the Master of Science in Data Analytics program at The Pennsylvania State University, Malvern, PA 19355 USA. Contact him at faizanraza766@gmail.com.

**Shuo (Luna) Yang** is an assistant professor of business at The Pennsylvania State University, Media, PA 19063 USA. Contact her at sfy5287@psu.edu.

**Satish Mahadevan Srinivasan** is an associate professor of information science at The Pennsylvania State University, Malvern, PA 19355 USA. Contact him at sus64@psu.edu.